\documentclass[manuscript,screen]{ACM/acmart}

\AtBeginDocument{%
  }

\acmYear{2026}
\setcopyright{acmlicensed}
\copyrightyear{2026}

\acmDOI{}

\usepackage{algorithm}
\usepackage{algorithmic}
\usepackage{placeins}
\usepackage{colortbl}
\usepackage{tabularx}

\definecolor{localmetric}{HTML}{E8F1F8}
\definecolor{globalmetric}{HTML}{F8EEE7}

\graphicspath{{figure/}{media/}}

\begin{document}

\title{From Graphs to Feeders: Constraint-Guided Diffusion for Rule-Compliant Feeder Generation}

\author{Yu Qin}
\email{yqin@nlr.gov}
\orcid{0003-0657-1149}
\correspondingauthor
\authornotemark[1]
\affiliation{%
  \institution{National Laboratory of the Rockies}
  \city{Golden}
  \state{Colorado}
  \country{USA}}

\author{Andrew Glaws}
\email{aglaws@nlr.gov}
\affiliation{%
  \institution{National Laboratory of the Rockies}
  \city{Golden}
  \state{Colorado}
  \country{USA}}

\author{Aadil Latif}
\email{alatif@nlr.gov}
\affiliation{%
  \institution{National Laboratory of the Rockies}
  \city{Golden}
  \state{Colorado}
  \country{USA}}

\author{Ryan King}
\email{rking@nlr.gov}
\affiliation{%
  \institution{National Laboratory of the Rockies}
  \city{Golden}
  \state{Colorado}
  \country{USA}}

\renewcommand{\shortauthors}{Qin et al.}

\begin{abstract}
Generative modeling approaches often focus on recovering broad statistical characteristics from the training data. In the context of graph generation, this may refer to degree distributions, clustering coefficients, or spectral properties. However, generating usable distribution feeders when detailed feeder models are unavailable requires more than matching generic graph statistics: the sampled topology must also obey electrical compatibility and radiality rules. We therefore formulate feeder synthesis as a constraint-guided graph generation problem and propose the Power-Grid-constrained Discrete Denoising Diffusion model, \textbf{PG-DiGress}, which learns categorical node and edge patterns from feeder data, while respecting domain-specific rules. Specifically, it injects feeder constraints into the reverse diffusion process through soft masks that suppress incompatible edge classes during denoising, followed by a final projection step that rebuilds a connected, rule-compliant feeder graph. 
We evaluate PG-DiGress using graph-distribution similarity, feeder-rule satisfaction, structural validity, and downstream model construction. Compared with the unconstrained baseline, PG-DiGress increases the strict feeder pass rate from $13.7\%$ to $96.8\%$. We also successfully convert the generated graphs into executable feeder models for downstream analysis.

\end{abstract}



\maketitle

\section{Introduction}
\label{sec:intro}
Distribution-feeder models are a basic input to many power-system studies. A feeder model describes how the source, primary conductors, transformers, secondary services, and loads are connected, together with the electrical attributes needed for analysis. Planning and operations studies use these models to evaluate quantities such as voltage profiles, equipment loading, losses, distributed-energy-resource integration, and the consequences of outages or infrastructure changes. Resilience and scenario studies similarly modify the network, loading, or operating conditions to examine how a feeder responds to hazards and alternative designs~\cite{Postigo_Marcos_Mateo_Gomez_Elgindy_Duenas_Palmintier_Hodge_Krackar_2017, Krishnan_Bugbee_Elgindy_Mateo_Duenas_Postigo_Lacroix_Gomez_Palmintier_2020}. In all of these cases, downstream analysis assumes that a sufficiently complete network model already exists.

In practice, feeder models are analyzed using distribution-system simulation and planning tools, such as OpenDSS~\cite{dugan2011reference} and GridLAB-D~\cite{chassin2008gridlabd}. These tools can solve power flow or time-series simulations once the buses, lines, transformers, phases, loads, and equipment parameters are specified, but they do not infer a missing network topology. Detailed utility feeder models are also often unavailable because of confidentiality, security, and data-access limitations, while public test systems capture only a small portion of the structural diversity found in real distribution networks~\cite{Postigo_Marcos_Mateo_Gomez_Elgindy_Duenas_Palmintier_Hodge_Krackar_2017, Palmintier_Elgindy_Mateo_Postigo_Gomez_deCuadra_Martinez_2021}. This data gap limits the construction of simulation models for studying alternative infrastructure designs, uncertain operating conditions, and future grid scenarios.

Engineering-based synthetic feeder tools provide one solution. They combine geographic information, distribution-design rules, optimization, and equipment libraries to construct realistic-but-not-real distribution networks~\cite{Palmintier_Elgindy_Mateo_Postigo_Gomez_deCuadra_Martinez_2021, Meyur_Vullikanti_Swarup_Mortveit_Centeno_Phadke_Poor_Marathe_2022, Duwadi_McKenna_Nagarajan_2022}. For example, roads and building locations can be used to estimate load placement and candidate routing, after which engineering heuristics determine transformer placement, conductor selection, and network connectivity. 
Learned graph generation offers a complementary capability: instead of
constructing a network from prescribed inputs and design assumptions, a generative model can learn recurring structural variation across a collection of feeder graphs and sample alternative candidate topologies. Distribution feeders are naturally represented as attributed graphs, and their node roles, phase labels, and connection types are categorical, making discrete graph diffusion a natural generative formulation~\cite{
Zhu_Du_Wang_Xu_Zhang_Liu_Wu_2022, Haefeli_Martinkus_Perraudin_Wattenhofer_2022, Vignac_Krawczuk_Siraudin_Wang_Cevher_Frossard_2023}.

However, learning a statistically plausible graph is not sufficient for feeder synthesis. Generic graph generators~\cite{You_Ying_Ren_Hamilton_Leskovec_2018, Simonovsky_Komodakis_2018, Bojchevski_Shchur_Zugner_Gunnemann_2018, Madhawa_Ishiguro_Nakago_Abe_2019, Vignac_Krawczuk_Siraudin_Wang_Cevher_Frossard_2023} do not explicitly encode the electrical meaning of their predicted node and edge classes. A conductor must connect compatible voltage classes and phases, a transformer must represent an appropriate primary--secondary transition, and the complete topology must provide source-connected radial paths to its loads. A graph can therefore match node count, degree, path-length, or spectral statistics while still containing incompatible conductors, misplaced transformers, disconnected loads, or multiple source-to-load paths. Such samples may look graph-like but cannot directly serve as usable feeder topologies.

\begin{figure}[ht]
    \centering
    \includegraphics[width=0.7\linewidth]{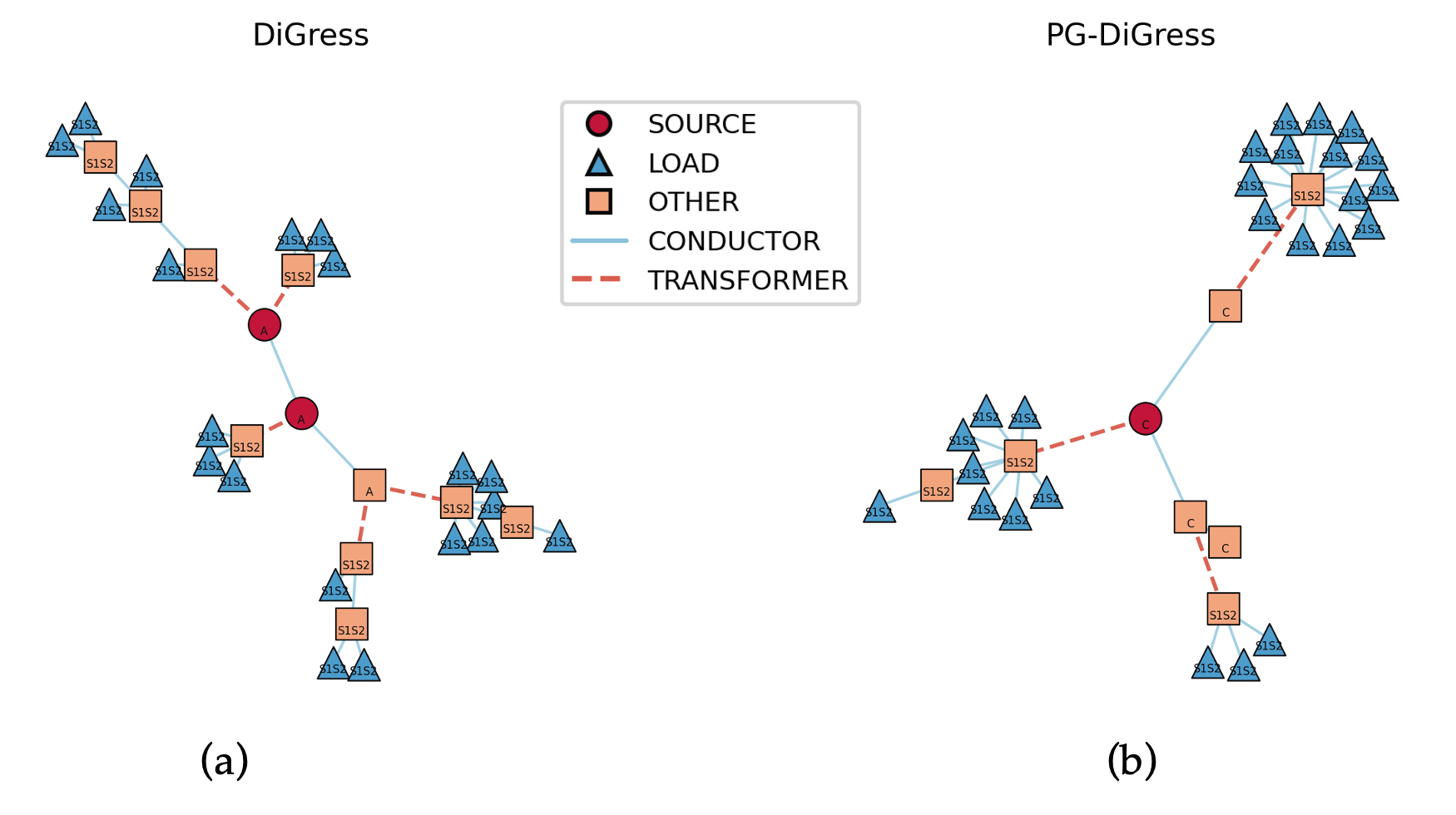}
    \Description{Side-by-side feeder graphs from unconstrained DiGress and PG-DiGress, highlighting the extra source in the unconstrained sample and the single-source rule-compliant structure in the PG-DiGress sample.}
    \caption{
    Representative feeder samples generated by (a) unconstrained DiGress and (b) PG-DiGress. The DiGress sample contains two source nodes and therefore violates the single-source feeder requirement despite exhibiting a plausible branching structure. In contrast, the PG-DiGress sample contains a single source and forms a source-connected topology with conductor and transformer assignments consistent with the sampled node attributes.
    }
    \label{fig:qual_exp}
\end{figure}

Figure~\ref{fig:qual_exp} (a) illustrates this distinction. The unconstrained graph generator (DiGress) sample exhibits a plausible branching structure but contains two source nodes, violating a basic feeder requirement. This example highlights
why graph-level plausibility alone is insufficient and motivates incorporating feeder knowledge directly into generation.

We therefore investigate whether discrete graph diffusion can learn data-driven feeder variation while explicit feeder rules improve the rule compliance and downstream usability of generated samples. We propose Power-Grid-constrained DiGress (\textbf{PG-DiGress}), a constraint-guided discrete diffusion framework
built on DiGress~\cite{Vignac_Krawczuk_Siraudin_Wang_Cevher_Frossard_2023}.
PG-DiGress learns categorical node and edge patterns from feeder data while introducing electrical rules into generation. During reverse diffusion, soft masks guide conductor and transformer predictions toward locally compatible assignments. After denoising, a final projection uses the learned edge probabilities to construct the required global feeder structure. 

Our contributions are:
\begin{itemize}
    \item We formulate distribution feeder synthesis as constraint-guided discrete graph generation with categorical node types, node phases, and edge classes.
    \item We introduce PG-DiGress, which combines soft masks during reverse diffusion to ensure local compatibility, and a final projection step to ensure global feeder validity.
    \item We evaluate generated feeders beyond standard graph statistics using domain-specific rule satisfaction, structural validity, and downstream feeder-model construction.
\end{itemize}

\section{Background and Motivation}
\label{sec:related}

We first describe how synthetic distribution feeders are constructed and validated in power-system studies. We then introduce learned graph generation and discrete graph diffusion, providing the background needed to understand where learning-based generation complements existing engineering approaches. We review the work most relevant to these foundations and refer readers to~\cite{Zhu_Du_Wang_Xu_Zhang_Liu_Wu_2022} for a broader survey of learned graph generation.

\subsection{Synthetic Power-System Generation}
Synthetic power-system modeling addresses the limited availability of shareable utility network data by constructing realistic-but-not-real systems. For distribution networks, existing workflows commonly begin with geographic and demand information, then apply engineering rules, optimization, and equipment catalogs to produce a complete feeder model. Meyur et al.~\cite{Meyur_Vullikanti_Swarup_Mortveit_Centeno_Phadke_Poor_Marathe_2022} generate geographically grounded distribution networks from roads, buildings, and engineering constraints. The Simple Synthetic Distribution Feeder Generation Tool (SHIFT)~\cite{Duwadi_McKenna_Nagarajan_2022} similarly constructs feeder models from OpenStreetMap data~\cite{Haklay_Weber_2008} and distribution-system design principles~\cite{Kersting_2012}. 

These engineering approaches provide an explicit construction process: geography and demand define where service is required, while design rules, optimization objectives, and equipment choices determine how that service is connected. Their outputs are therefore interpretable and can often be parameterized directly for downstream simulation. Learned generation addresses a different question considered in this work: whether recurring structural
patterns across an existing feeder population can themselves be represented
as a probability distribution and sampled to produce alternative candidate topologies.

Synthetic-system validation also requires more than checking graph shape. Krishnan et al.~\cite{Krishnan_Bugbee_Elgindy_Mateo_Duenas_Postigo_Lacroix_Gomez_Palmintier_2020} evaluate synthetic distribution systems using physical layout, component and topology statistics, power-flow behavior, voltage profiles, losses, and expert criteria. This distinction is important for our evaluation: statistical similarity measures agreement with the reference feeder population, whereas feeder-rule compliance measures whether the categorical topology satisfies the electrical and structural relationships encoded in our representation.

Recent work has also begun to explore learned power-grid synthesis. PowerGrow~\cite{He_Xiao_Li_Qiu_Xu_Weng_He_Tong_2025}, for example, jointly models aspects of topology, continuous attributes, and temporal behavior. Our focus is narrower but complementary: we study whether categorical distribution-feeder topology can be learned from data while explicit phase, voltage-class, conductor, transformer, and radiality rules are incorporated directly into generation.

\subsection{Learned Graph Generation}
A graph neural network operates on graph-structured data by updating node or edge representations using information from neighboring elements and is commonly used for prediction on an existing graph~\cite{battaglia2018relational}. Graph generation poses a different problem: the model must generate the graph itself, including its connectivity and attributes. Learned graph generators therefore model a probability distribution over graph-structured objects and sample new graphs from that distribution. Graphs are more difficult to generate than fixed-size vectors or images because node ordering is not unique, graph size can vary, and local edge decisions collectively determine global properties such as connectivity, cycles, paths, and components~\cite{
You_Ying_Ren_Hamilton_Leskovec_2018, Liao_Li_Song_Wang_Nash_Hamilton_Duvenaud_Urtasun_Zemel_2019, Jo_Lee_Hwang_2022}.

Prior approaches include autoregressive, latent-variable, adversarial, normalizing-flow, and diffusion-based models. Autoregressive methods generate nodes and edges sequentially, providing explicit dependence on the partial graph but requiring many ordered decisions~\cite{
You_Ying_Ren_Hamilton_Leskovec_2018, Liao_Li_Song_Wang_Nash_Hamilton_Duvenaud_Urtasun_Zemel_2019}. Latent-variable, adversarial, and flow-based methods generate larger portions of a graph jointly, but may face limitations related to graph matching, scalability, indirect topology construction, or invertibility~\cite{
Simonovsky_Komodakis_2018, Bojchevski_Shchur_Zugner_Gunnemann_2018, Madhawa_Ishiguro_Nakago_Abe_2019}. These methods differ in how they represent and sample the joint dependencies among nodes and edges. For feeder synthesis, these dependencies also carry explicit electrical
semantics: whether an edge class is admissible depends on the voltage class
and phase configuration of its endpoints. This motivates incorporating domain knowledge into the generation process rather than evaluating such relationships only after sampling.

\subsection{Discrete Graph Diffusion}
Diffusion models formulate generation as the reversal of a known corruption process~\cite{ho2020denoising}. In the forward process, a training sample is gradually transformed toward noise. A denoising network is trained at intermediate noise levels to predict the clean sample or the reverse transition. During generation, the model begins with a noisy state and repeatedly applies the learned reverse transitions to obtain a new sample. 
For graphs with categorical attributes, discrete diffusion performs this corruption directly in the categorical state rather than perturbing a continuous adjacency representation with Gaussian noise~\cite{Haefeli_Martinkus_Perraudin_Wattenhofer_2022}. This formulation is particularly suitable for feeder representation: node classes (\texttt{SOURCE}, \texttt{LOAD}), node phases, and edge classes (\texttt{CONDUCTOR}, \texttt{TRANSFORMER}) are discrete semantic categories rather than continuous measurements.

Discrete Denoising Diffusion for Graph Generation (DiGress)~\cite{Vignac_Krawczuk_Siraudin_Wang_Cevher_Frossard_2023} implements this idea by applying categorical Markov transitions to node and edge labels and training a graph transformer to reconstruct their clean states. Subsequent work improves scalability or conditioning ~\cite{ Chen_He_Han_Liu_2023, Qin_Vignac_Frossard_2023, Tseng_Diamant_Biancalani_Scalia_2023}, but the conditions are generally expressed through learned or generic graph properties rather than explicit feeder-specific dependencies between node attributes and electrical edge classes.

PG-DiGress therefore retains discrete diffusion as the mechanism for learning
feeder variation while introducing feeder rules into sampling. The following section formalizes the categorical feeder representation and the local and global rules used to guide generation.

\begin{figure}[!t]
    \centering
    \begin{minipage}[t]{0.42\linewidth}
        \centering
        \includegraphics[width=\linewidth]{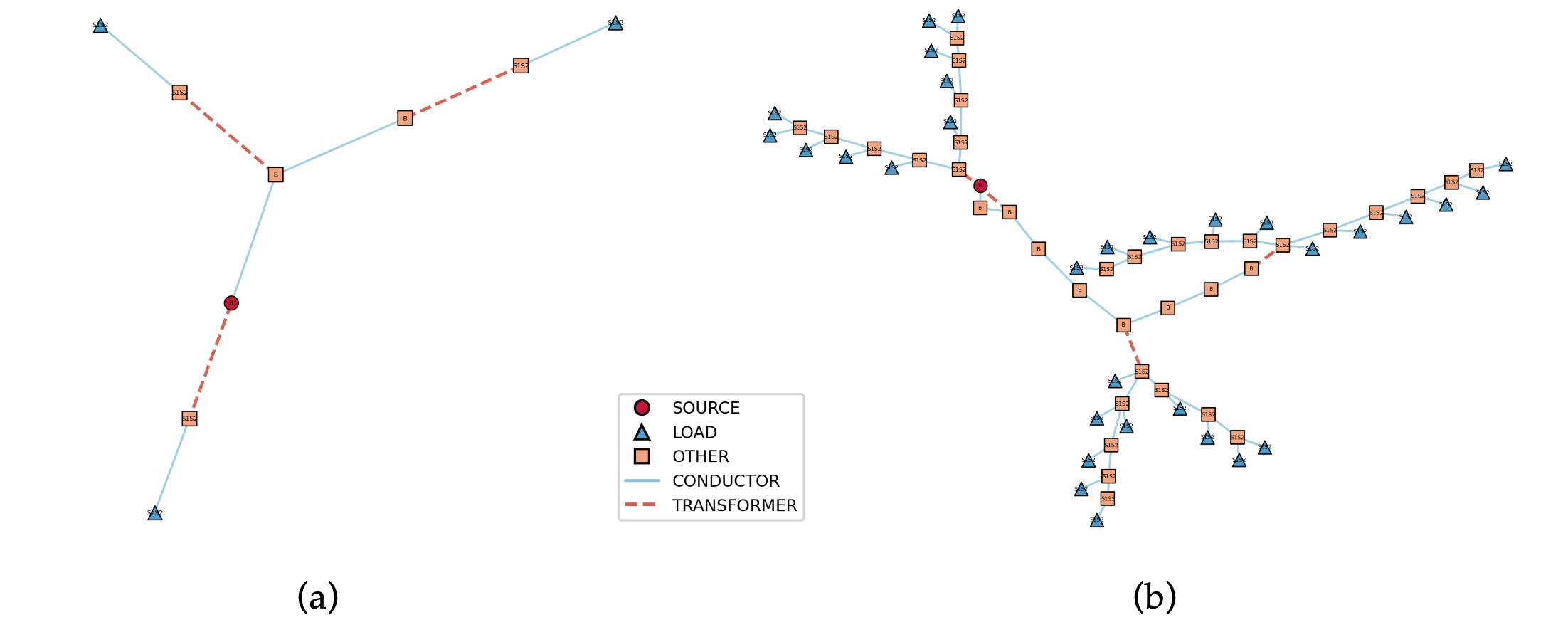}
    \end{minipage}
    \hfill
    \begin{minipage}[t]{0.56\linewidth}
        \centering
        \includegraphics[width=\linewidth]{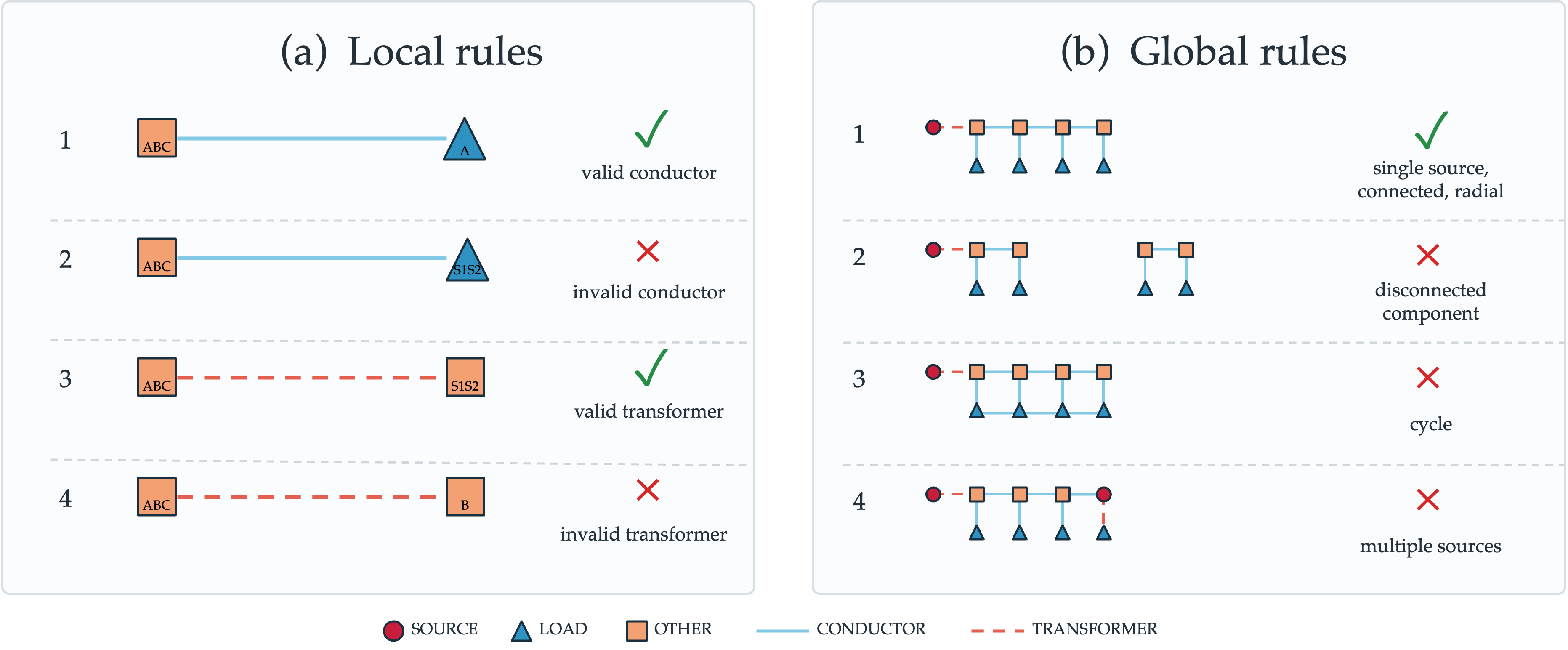}
    \end{minipage}
    \Description{Two-panel figure showing representative attributed SMART-DS feeder graphs and the local edge-compatibility and global topology rules used for rule-compliant generation.}

    \caption{
    Feeder representation and rule system used in this work.
    Left: representative attributed SMART-DS feeder graphs, with nodes
    encoding role and phase and edges encoding conductor or transformer
    connections. Right: local edge-compatibility rules and representative
    global topology requirements used to define rule-compliant generation.
    }
    \label{fig:problem_setup}
\end{figure}

\section{Problem Formulation}
\label{sec:problem}
We formulate distribution-feeder synthesis as generation over attributed graphs subject to explicit structural and electrical compatibility rules. The representation specifies the node and edge attributes learned by the generative model, while the rule system identifies which generated graphs correspond to usable distribution feeders. Figure~\ref{fig:problem_setup} summarizes the feeder representation and the corresponding local and global rule system.

\subsection{Feeder Graph Representation}

We represent a power distribution feeder as an undirected attributed graph for modeling purposes: conductor and transformer connectivity is symmetric, whereas power-flow direction depends on the operating state and may change with distributed generation. Once a single-source radial topology is generated, it can be rooted at the source to recover a parent--child orientation for downstream analysis. Thus, we define the feeder graph:
\begin{equation}
    G = (V,X,P,E),
    \label{eq:feeder-representation}
\end{equation}
where $V$ is the node set, 
$X=\{X_v\}_{v\in V}$ is the categorical type of node,
$P=\{P_v\}_{v\in V}$ is its categorical phase label, and 
$E=\{E_{uv}\}_{u,v\in V}$ is the categorical edge labels. Specifically, $X_v\in\mathcal{X}$, $P_v\in\mathcal{P}$, and $E_{uv}\in\mathcal{L}_E$. Because the feeder is modeled as undirected,
$E_{uv}=E_{vu}$. The attribute spaces are application-dependent and can be adapted to other feeder datasets without changing the underlying generative formulation.

For the dataset used in this work, the categorical spaces are
\begin{equation}
\begin{aligned}
    \mathcal{X}
    &=
    \{
    \texttt{SOURCE},
    \texttt{LOAD},
    \texttt{OTHER}
    \}, \\
    \mathcal{P}_{\mathrm{pri}}
    &=
    \{
    \texttt{ABC},
    \texttt{A},
    \texttt{B},
    \texttt{C},
    \texttt{AB},
    \texttt{BC},
    \texttt{AC}
    \}, \\
    \mathcal{P}_{\mathrm{sec}}
    &=
    \{
    \texttt{S1},
    \texttt{S2},
    \texttt{S1S2},
    \texttt{NS1S2}
    \}, \\
    \mathcal{P}
    &=
    \mathcal{P}_{\mathrm{pri}}
    \cup
    \mathcal{P}_{\mathrm{sec}}, \\
    \mathcal{L}_E
    &=
    \{
    \texttt{NO\_EDGE},
    \texttt{CONDUCTOR},
    \texttt{TRANSFORMER}
    \}.
\end{aligned}
\label{eq:dataset-label-spaces}
\end{equation}
Here, \texttt{SOURCE} denotes the feeder head, \texttt{LOAD} denotes an end-use load, and \texttt{OTHER} denotes an intermediate component such as a pole or bus. The edge label \texttt{NO\_EDGE} is included as a discrete class so that graph generation can be formulated as categorical prediction over all candidate node pairs.

Although node type and phase are written separately for semantic clarity, PG-DiGress represents them jointly through
\begin{equation}
    Z_v=(X_v,P_v)\in\mathcal{Z},
    \qquad
    \mathcal{Z}\subseteq\mathcal{X}\times\mathcal{P},
    \label{eq:joint-node-space}
\end{equation}
where $\mathcal{Z}$ contains the admissible type--phase combinations in the processed feeder representation. In particular, source labels are restricted to primary-side phase classes:
\begin{equation}
    (\texttt{SOURCE},p)\in\mathcal{Z}
    \quad\Rightarrow\quad
    p\in\mathcal{P}_{\mathrm{pri}}.
    \label{eq:source-primary-admissibility}
\end{equation}
Thus, source--primary consistency holds by construction. In contrast,
\texttt{LOAD} is a semantic node role and may occur on either the primary or secondary side of the feeder.

Each phase label is interpreted through a mapping $\psi:\mathcal{P}\rightarrow 2^{\Phi}$, where $\Phi$ is the primitive phase alphabet. For example, $\psi(\texttt{ABC})=\{\texttt{A},\texttt{B},\texttt{C}\}$ and $\psi(\texttt{S1S2})=\{\texttt{S1},\texttt{S2}\}$. This interpretation allows phase compatibility to be evaluated through set intersection.

The phase label also determines the voltage class through
\begin{equation}
    \nu(P_v)
    =
    \begin{cases}
        \texttt{PRIMARY},
        & P_v \in \mathcal{P}_{\mathrm{pri}},\\
        \texttt{SECONDARY},
        & P_v \in \mathcal{P}_{\mathrm{sec}}.
    \end{cases}
    \label{eq:voltage-class}
\end{equation}
We write $\nu_v:=\nu(P_v)$. Thus, \texttt{PRIMARY} and \texttt{SECONDARY} are derived electrical classes rather than additional generated node types.

Because $E$ assigns a categorical label to every node pair, we distinguish the edge-label tensor from the realized feeder topology. The feeder is undirected, $E_{uv}=E_{vu}$. We define the set of physical edges as
\begin{equation}
    \mathcal{E}_{\mathrm{phys}}(E)
    =
    \left\{
    \{u,v\}
    \;\middle|\;
    u,v\in V,\;
    u<v,\;
    E_{uv}\neq\texttt{NO\_EDGE}
    \right\},
    \qquad
    G_{\mathrm{phys}}(E)=\left(V,\mathcal{E}_{\mathrm{phys}}(E)\right).
    \label{eq:physical-graph}
\end{equation}
Thus, $G_{\mathrm{phys}}$ contains only conductor and transformer edges and is the graph for evaluation. 

\subsection{Feeder Rule System}

Using the representation above, distribution-system knowledge can be written as constraints on $X$, $P$, $E$, and the topology of $G_{\mathrm{phys}}$. We distinguish local rules, which determine whether a particular node or edge assignment is compatible, and global rules, which determine whether the complete graph has a valid feeder organization. These constraints establish structural feeder validity but do not replace downstream power-flow analysis of voltage, current, loading, or losses.

\paragraph{Local compatibility rules.}

In a common distribution-system dataset, the source belongs to the primary system, while load nodes may appear on either the primary or secondary system depending on their phase label. A conductor must remain within one voltage class and carry at least one phase shared by its endpoints, whereas a transformer connects the primary and secondary systems. Self-loops are not physically meaningful. These conditions are summarized as
\begin{equation}
\begin{aligned}
    X_v=\texttt{SOURCE}
    &\Rightarrow
    \nu_v=\texttt{PRIMARY},\\
    E_{uv}=\texttt{CONDUCTOR}
    &\Rightarrow
    \nu_u=\nu_v
    \;\land\;
    \psi(P_u)\cap\psi(P_v)\neq\emptyset,\\
    E_{uv}=\texttt{TRANSFORMER}
    &\Rightarrow
    \nu_u\neq\nu_v,\\
    E_{vv}
    &=\texttt{NO\_EDGE},
    \qquad \forall v\in V.
\end{aligned}
\label{eq:local-feeder-rules}
\end{equation}
Node role and voltage class are distinct attributes in our representation.
In particular, a \texttt{LOAD} node may occur on either the primary or
secondary side of the feeder and its voltage class is determined by its phase
label through $\nu(P_v)$. The source, however, belongs to the primary system.
The conductor rule reflects that a line does not change nominal voltage and can carry power only through phases available at both endpoints. The transformer rule captures the voltage-class transition between the primary feeder and secondary service network. Together, these rules enforce essential electrical compatibility conditions. 

\paragraph{Global topology rules.}

We define the source, load, and primary node sets as
\begin{equation}
\begin{aligned}
    V_{\mathrm{src}}
    &=
    \{v\in V\mid X_v=\texttt{SOURCE}\},\\
    V_{\mathrm{load}}
    &=
    \{v\in V\mid X_v=\texttt{LOAD}\},\\
    V_{\mathrm{pri}}
    &=
    \{v\in V\mid \nu_v=\texttt{PRIMARY}\}.
\end{aligned}
\end{equation}
The primary conductor subgraph is $G_{\mathrm{pri}}=(V_{\mathrm{pri}},\mathcal{E}_{\mathrm{pri}})$, where
\begin{equation}
    \mathcal{E}_{\mathrm{pri}}
    =
    \left\{
    \{u,v\}\in\mathcal{E}_{\mathrm{phys}}(E)
    \;\middle|\;
    u,v\in V_{\mathrm{pri}},
    \;
    E_{uv}=\texttt{CONDUCTOR}
    \right\}.
\end{equation}

Let $\Pi_G(v,s)$ denote the set of simple paths between node $v$ and source $s$ in $G_{\mathrm{phys}}(E)$. For a path $\pi$, define
\begin{equation}
    N_{\mathrm{tr}}(\pi)
    =
    \sum_{\{i,j\}\in\pi}
    \mathbf{1}
    \left[
    E_{ij}=\texttt{TRANSFORMER}
    \right]
\end{equation}
as the number of transformer edges on path $\pi$. The feeder topology must satisfy
\begin{equation}
\begin{aligned}
    |V_{\mathrm{src}}|
    &=1,
    \qquad V_{\mathrm{src}}=\{s\},\\
    G_{\mathrm{phys}}(E)
    &\text{ is connected},\\
    G_{\mathrm{pri}}
    &\text{ is a tree},\\
    \forall v\in V_{\mathrm{load}}:\qquad
    \Pi_G(v,s)
    &=\{\pi_{v,s}\},\\
    N_{\mathrm{tr}}(\pi_{v,s})
    &=
    \mathbf{1}
    \left[
        \nu_v=\texttt{SECONDARY}
    \right].
\end{aligned}
\label{eq:global-feeder-rules}
\end{equation}

The unique-source and connectivity requirements ensure that every modeled
component belongs to a source-connected feeder. Requiring the primary
conductor subgraph to be a tree enforces the radial primary structure
considered in this work. Finally, every load must have a unique path to the
source. A primary-side load is reached without crossing a transformer,
whereas a secondary-side load crosses exactly one transformer, corresponding
to a single primary-to-secondary voltage transition.

\subsection{Generation Objective}

The graph representation defines the ambient space $\mathcal{G}$ of attributed feeder graphs. The local compatibility and global topology rules in Eqs.~\eqref{eq:local-feeder-rules}--\eqref{eq:global-feeder-rules} define the rule-compliant subset
\begin{equation}
    \mathcal{G}_{\mathcal{R}}
    =
    \left\{
    G\in\mathcal{G}
    \;\middle|\;
    G\text{ satisfies all rules in }\mathcal{R}
    \right\}.
    \label{eq:feasible-feeder-space}
\end{equation}
where $\mathcal{R}$ denotes the complete feeder rule system.

Given a training set $\mathcal{D}=\{G_i\}_{i=1}^{N}$ sampled from an unknown feeder distribution, our goal is to learn a rule-guided generative distribution $p_{\theta}^{\mathcal{R}}$ such that
\begin{equation}
    \hat{G}\sim p_{\theta}^{\mathcal{R}}(G),
    \qquad
    p_{\theta}^{\mathcal{R}}
    \left(\mathcal{G}_{\mathcal{R}}\right)
    \approx 1.
    \label{eq:generation-objective}
\end{equation}
Here, the superscript $\mathcal{R}$ indicates that the learned sampling process is modified by the feeder rules. In addition to rule compliance, samples drawn from $p_{\theta}^{\mathcal{R}}$ should preserve structural variation observed in $\mathcal{D}$ and generate graphs that can be converted into downstream feeder models.

Sampling directly from $\mathcal{G}_{\mathcal{R}}$ is challenging because local node--edge compatibility and global feeder topology are coupled. The next section introduces our approach, which targets this objective through local edge guidance during reverse diffusion and an edge-only projection after denoising.

\section{Methodology}
\label{sec:method}

\begin{sloppypar}
We propose Power-Grid-constrained Discrete Denoising Diffusion (PG-DiGress) to generate distribution-feeder topologies that preserve data-driven structural variation while better satisfying feeder rules. PG-DiGress follows a \emph{minimal-intervention} design. We retain the
standard DiGress training objective and diffusion process, and introduce feeder knowledge only during edge sampling and final topology construction. This keeps training stable and makes the effect of rule guidance directly comparable with the unconstrained model.
\end{sloppypar}

\begin{figure}[t]
    \centering
    \includegraphics[width=\linewidth]{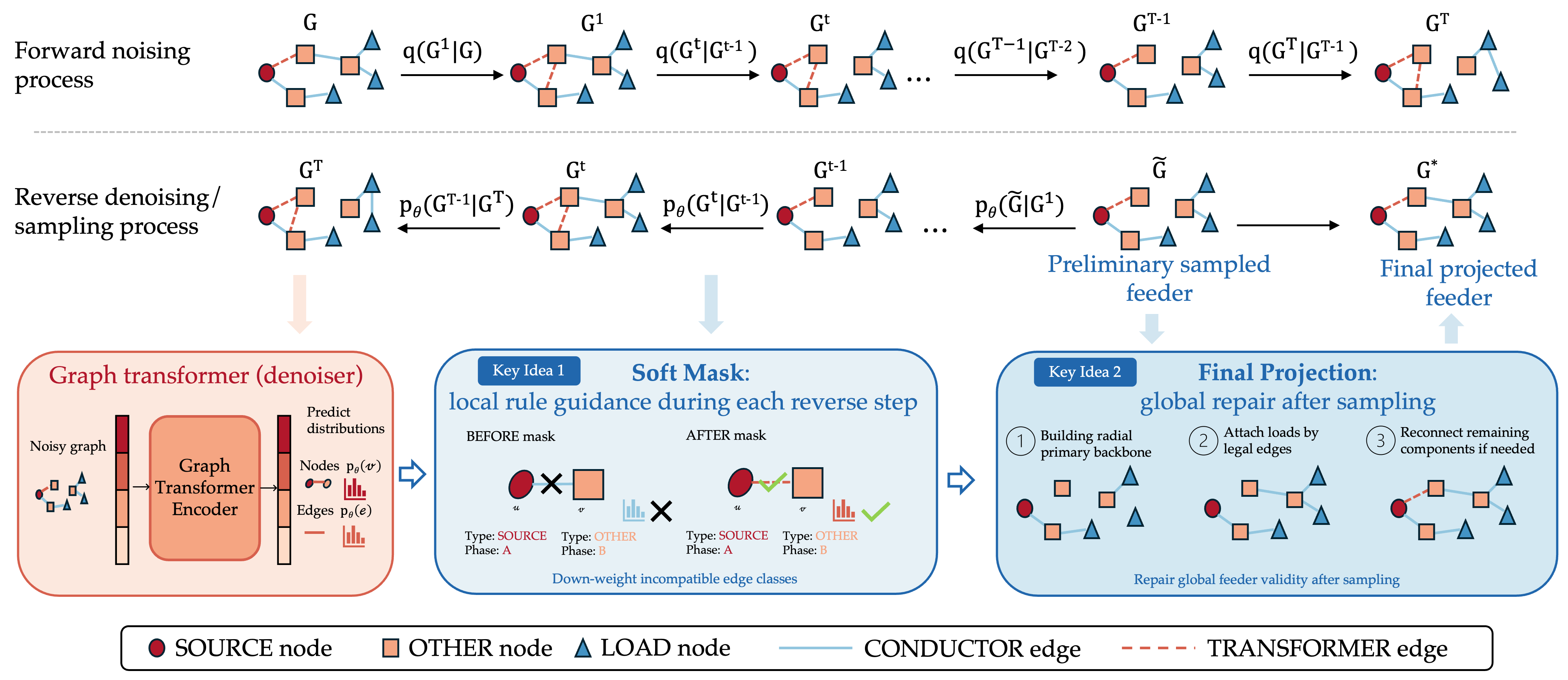}
    \Description{PG-DiGress reverse diffusion pipeline with graph-transformer predictions, a local compatibility mask for edge classes, and a final projection that builds the global feeder structure.}
    \caption{
    Overview of PG-DiGress. At each reverse diffusion step, a graph transformer (denoiser) predicts categorical node and edge distributions. The predicted node types and phases define a soft mask that re-weights conductor and transformer classes according to local compatibility rules. After denoising, a final projection uses the predicted edge probabilities to build a radial primary backbone, attaches loads, and reconnects remaining components while keeping the sampled node labels fixed.
    }
    \label{fig:overview}
\end{figure}

The method follows the distinction established in the previous section: local electrical compatibility can be incorporated during reverse diffusion, whereas global topology requirements must be addressed at the graph level after denoising. Accordingly, PG-DiGress contains two main components: a \emph{soft mask} that guides each reverse diffusion step and a \emph{final projection} that repairs the final topology. Figure~\ref{fig:overview} shows the generation pipeline.

\subsection{Discrete Graph Diffusion}

PG-DiGress builds on Discrete Denoising Diffusion for Graph Generation
(DiGress)~\cite{Vignac_Krawczuk_Siraudin_Wang_Cevher_Frossard_2023}, which models graphs with categorical node and edge attributes. We summarize the components needed to describe our method and refer readers to the original work for the complete diffusion formulation. 

Following Eq.~\eqref{eq:joint-node-space}, the implementation jointly encodes
node type and phase as the categorical node label
\begin{equation}
    Z_v=(X_v,P_v)\in\mathcal{Z}.
\end{equation}
The semantic attributes are recovered through deterministic decoding functions
\begin{equation}
    X_v=d_X(Z_v),
    \qquad
    P_v=d_P(Z_v).
    \label{eq:node-label-decoding}
\end{equation}

At diffusion step $t$, the graph is represented as
\begin{equation}
    G^t=(V,Z^t,E^t),
    \qquad
    t=0,\ldots,T,
\end{equation}
where $G^0$ is a clean feeder graph and $G^T$ approaches a categorical noise distribution. The node set $V$ remains fixed throughout the diffusion trajectory, while the node labels $Z^t$ and edge labels $E^t$ are progressively corrupted.

\paragraph{Forward Diffusion}

The forward process independently applies discrete Markov transitions to node and edge labels:
\begin{equation}
\begin{aligned}
    q(G^t\mid G^{t-1})
    &=
    \prod_{v\in V}
    \operatorname{Cat}
    \left(
    Z_v^t;
    Z_v^{t-1}Q_Z^t
    \right) \\
    &\quad\times
    \prod_{u<v}
    \operatorname{Cat}
    \left(
    E_{uv}^t;
    E_{uv}^{t-1}Q_E^t
    \right),
\end{aligned}
\label{eq:forward-diffusion}
\end{equation}
where $Q_Z^t$ and $Q_E^t$ are transition matrices for node and edge labels. Because the feeder graph is undirected, edge labels are sampled only for $u<v$ and then symmetrized.

\paragraph{Denoising Model}

A graph transformer $\phi_\theta$ receives the noisy graph $G^t$ and timestep $t$ and predicts categorical distributions over the corresponding clean labels:
\begin{equation}
    \phi_\theta(G^t,t)
    =
    \left(
        \left\{
        \hat{p}_{Z,v,\theta}^{0}
        (\cdot\mid G^t,t)
        \right\}_{v\in V},
        \left\{
        \hat{p}_{E,uv,\theta}^{0}
        (\cdot\mid G^t,t)
        \right\}_{u<v}
    \right).
    \label{eq:denoiser-output}
\end{equation}

The denoiser is trained using categorical cross-entropy:
\begin{equation}
\begin{aligned}
    \mathcal{L}_{\mathrm{diff}}
    &=
    -\sum_{v\in V}
    \log
    \hat{p}_{Z,\theta}^{0}
    \left(
    Z_v^0\mid G^t,t
    \right) \\
    &\quad
    -\lambda_E
    \sum_{u<v}
    \log
    \hat{p}_{E,\theta}^{0}
    \left(
    E_{uv}^0\mid G^t,t
    \right),
\end{aligned}
\label{eq:diffusion-loss}
\end{equation}
where $\lambda_E$ balances node- and edge-label prediction.

Following DiGress, the predicted clean-label distributions are combined with the closed-form categorical diffusion posterior to obtain reverse-step probabilities. We denote the resulting node and edge probabilities by
\begin{equation}
\begin{aligned}
    \pi_{Z,v}^t(z)
    &:=
    p_\theta
    \left(
    Z_v^{t-1}=z
    \mid
    G^t
    \right),\\
    \pi_{E,uv}^t(k)
    &:=
    p_\theta
    \left(
    E_{uv}^{t-1}=k
    \mid
    G^t
    \right).
\end{aligned}
\label{eq:baseline-reverse-probabilities}
\end{equation}

The baseline reverse distribution factorizes as
\begin{equation}
\begin{aligned}
    p_\theta(G^{t-1}\mid G^t)
    &=
    \prod_{v\in V}
    \operatorname{Cat}
    \left(
    Z_v^{t-1};
    \pi_{Z,v}^t
    \right) \\
    &\quad\times
    \prod_{u<v}
    \operatorname{Cat}
    \left(
    E_{uv}^{t-1};
    \pi_{E,uv}^t
    \right).
\end{aligned}
\label{eq:baseline-reverse-distribution}
\end{equation}

Original DiGress samples both node and edge labels directly from Eq.~\eqref{eq:baseline-reverse-distribution}. Although this provides a natural generator for discrete feeder attributes, it does not explicitly account for electrical edge compatibility or global requirements such as source connectivity and radiality.

\subsection{PG-DiGress: Rule-Guided Sampling}
\label{sec:rule-guided-sampling}

PG-DiGress preserves the trained DiGress denoiser and modifies only the edge generation. We modify edge generation because node attributes define the electrical meaning of a candidate connection, while edge labels determine whether that connection is compatible. This provides a direct point for incorporating feeder rules without redesigning the node generator or retraining the denoiser. Its sampling process can be summarized as
\begin{equation}
    G^T
    \xrightarrow[\text{soft mask}]
    {\text{reverse diffusion}}
    \widetilde{G}
    \xrightarrow[\text{edge-only repair}]
    {\text{final projection}}
    G^*,
    \label{eq:pgdigress-flow}
\end{equation}
where $\widetilde{G}$ is the preliminary graph produced by guided reverse diffusion and $G^*$ is the final projected graph.

\subsection{Soft Mask for Local Edge Compatibility}

At every reverse step, the model first predicts the clean categorical node distribution. The most likely node label is decoded as
\begin{equation}
\begin{aligned}
    \bar{Z}_v^t
    &=
    \arg\max_{z\in\mathcal{Z}}
    \hat{p}_{Z,\theta}^{0}
    \left(
    z\mid G^t,t
    \right),\\
    \bar{X}_v^t
    &=
    d_X(\bar{Z}_v^t),\\
    \bar{P}_v^t
    &=
    d_P(\bar{Z}_v^t),\\
    \bar{\nu}_v^t
    &=
    \nu(\bar{P}_v^t).
\end{aligned}
\label{eq:decoded-node-metadata}
\end{equation}

Here, $\bar{X}_v^t$, $\bar{P}_v^t$, and $\bar{\nu}_v^t$ are temporary metadata used to evaluate edge compatibility. They do not replace the node labels sampled from $\pi_{Z,v}^t$.

For every node pair $\{u,v\}$ and edge class $k\in\mathcal{L}_E$, we define a binary legality mask
\begin{equation}
    M_{uv}^t(k)\in\{0,1\}.
\end{equation}
For distinct node pairs, \texttt{NO\_EDGE} is always admissible: 
\begin{equation}
    M_{uv}^t(\texttt{NO\_EDGE}) =1.
\end{equation}The mask encodes the two local rules from Eq.~\eqref{eq:local-feeder-rules}: conductors require equal voltage class and overlapping phases, whereas transformers connect the primary and secondary systems.
\begin{equation}
\begin{aligned}
    M_{uv}^t(\texttt{CONDUCTOR})
    &=
    \mathbf{1}
    \left[
        u\neq v
        \;\land\;
        \bar{\nu}_u^t=\bar{\nu}_v^t
        \;\land\;
        \psi(\bar{P}_u^t)
        \cap
        \psi(\bar{P}_v^t)
        \neq\emptyset
    \right],\\
    M_{uv}^t(\texttt{TRANSFORMER})
    &=
    \mathbf{1}
    \left[
    u\neq v
    \land
    \bar{\nu}_u^t\neq\bar{\nu}_v^t
    \right].
\end{aligned}
\label{eq:edge-legality-mask}
\end{equation}
Diagonal entries are fixed to \texttt{NO\_EDGE}, so self-loops cannot be sampled.

The mask reweights the baseline reverse edge probabilities. We define
\begin{equation}
    a_{uv}^t(k)
    =
    \exp
    \left(
        -\lambda_t
        \left[
        1-M_{uv}^t(k)
        \right]
    \right),
    \label{eq:soft-mask-weight}
\end{equation}
where $\lambda_t\geq0$ controls the guidance strength. Legal classes receive weight $1$, whereas illegal classes receive weight $\exp(-\lambda_t)$. Thus, $\lambda_t=0$ recovers the unconstrained DiGress distribution, while larger values increasingly suppress illegal classes. 

The guided reverse edge distribution is
\begin{equation}
    \widetilde{\pi}_{E,uv}^t(k)
    =
    \frac{
        \pi_{E,uv}^t(k)
        a_{uv}^t(k)
    }{
        \displaystyle
        \sum_{k'\in\mathcal{L}_E}
        \pi_{E,uv}^t(k')
        a_{uv}^t(k')
    }.
    \label{eq:masked-edge-distribution}
\end{equation}

Node labels are sampled from the original reverse node distribution:
\begin{equation}
    Z_v^{t-1}
    \sim
    \operatorname{Cat}
    \left(
    \pi_{Z,v}^t
    \right),
    \label{eq:node-sampling}
\end{equation}
while edge labels are sampled from the masked distribution:
\begin{equation}
    E_{uv}^{t-1}
    \sim
    \operatorname{Cat}
    \left(
    \widetilde{\pi}_{E,uv}^t
    \right).
    \label{eq:masked-edge-sampling}
\end{equation}

The soft mask therefore directly modifies the reverse distribution over edge sampling conditional on the current node predictions without altering the learned node generator because node predictions remain uncertain during denoising. This preserves probabilistic sampling while allowing the guidance strength to increase as predictions become more reliable, and the guidance can still indirectly influence later node predictions.

\subsection{Final Projection for Global Feeder Structure}

Pairwise masking improves local edge compatibility but cannot enforce the global feeder requirements introduced in Section~\ref{sec:problem}: a single source-connected feeder, a radial primary structure, and a unique source-to-load path with the voltage-class transition required by the destination load. These properties depend jointly on multiple edge decisions and therefore cannot be guaranteed by independently masking candidate node pairs. Enforcing them at every reverse step would require solving a global constrained graph problem throughout denoising. PG-DiGress instead addresses them through a final projection after the last reverse step.

Let
\begin{equation}
    \widetilde{G}
    =
    (V,\widetilde{Z},\widetilde{E})
\end{equation}
denote the preliminary sampled graph. The projection keeps the node labels fixed and modifies only the edge-label tensor:
\begin{equation}
\begin{aligned}
    E^*
    &=
    \Pi_{\mathrm{edge}}
    \left(
        \widetilde{E};
        \widetilde{Z},
        \rho,
        \mathcal{R}
    \right),\\
    G^*
    &=
    (V,\widetilde{Z},E^*),
\end{aligned}
\label{eq:edge-only-projection}
\end{equation}
where
\begin{equation}
    \rho_{uv}(k)
    :=
    \widetilde{\pi}_{E,uv}^{1}(k)
\end{equation}
is the final guided probability for edge class $k$. 

Because the masks used during diffusion are based on temporary node predictions, the projection recomputes edge compatibility from the final sampled labels $\widetilde{Z}$. We denote this final mask by
\begin{equation}
    M_{uv}^{*}(k)
    :=
    M
    \left(
        k;
        d_X(\widetilde{Z}_u),
        d_P(\widetilde{Z}_u),
        d_X(\widetilde{Z}_v),
        d_P(\widetilde{Z}_v)
    \right),
    \label{eq:final-edge-mask}
\end{equation}
using the same local rules as Eq.~\eqref{eq:edge-legality-mask}.

The final projection consists of three stages: the primary backbone construction, load attachment, and component reconnection.

\paragraph{Primary Backbone Construction}

The sampled node labels determine the primary-side node set
\begin{equation}
    V_{\mathrm{pri}}^{*}
    =
    \left\{
        v\in V
        \;\middle|\;
        \nu\left(d_P(\widetilde{Z}_v)\right)
        =
        \texttt{PRIMARY}
    \right\}.
    \label{eq:projected-primary-nodes}
\end{equation}

The legal primary conductor candidates are
\begin{equation}
    \mathcal{C}_{\mathrm{pri}}
    =
    \left\{
        \{u,v\}
        \;\middle|\;
        u,v\in V_{\mathrm{pri}}^{*},
        \;
        M_{uv}^{*}(\texttt{CONDUCTOR})=1
    \right\},
\end{equation}
with weights
\begin{equation}
    w_{uv}^{\mathrm{cond}}
    =
    \rho_{uv}(\texttt{CONDUCTOR}).
\end{equation}

We construct a maximum-weight spanning forest
\begin{equation}
    F_{\mathrm{pri}}
    =
    \operatorname{MaxSF}
    \left(
        V_{\mathrm{pri}}^{*},
        \mathcal{C}_{\mathrm{pri}},
        w^{\mathrm{cond}}
    \right).
    \label{eq:primary-backbone}
\end{equation}
When the legal candidate graph is connected, $F_{\mathrm{pri}}$ is a maximum-weight spanning tree. Otherwise, it provides an acyclic set of high-probability conductor edges, and the remaining components are handled by the reconnection stage.

\paragraph{Load Attachment}

The sampled load-node set is
\begin{equation}
    V_{\mathrm{load}}^{*}
    =
    \left\{
        v\in V
        \;\middle|\;
        d_X(\widetilde{Z}_v)
        =
        \texttt{LOAD}
    \right\}.
\end{equation}

Let $H$ denote the partially constructed feeder, initialized with the primary backbone. For each unattached load $v$, let $\mathcal{A}_H(v)$ denote the set of candidate parent and edge-type pairs $(u,k)$ such that:

\begin{itemize}
    \item $u$ already belongs to the source-connected component of $H$;
    \item $k\in\{\texttt{CONDUCTOR},\texttt{TRANSFORMER}\}$;
    \item $M_{uv}^{*}(k)=1$; and
    \item adding edge $(u,v)$ with label $k$ preserves the required source-to-load transformer structure.
\end{itemize}

When $\mathcal{A}_H(v)$ is nonempty, PG-DiGress selects
\begin{equation}
    (u^*(v),k^*(v))
    =
    \arg\max_{(u,k)\in\mathcal{A}_H(v)}
    \rho_{uv}(k)
    \label{eq:load-attachment}
\end{equation}
and adds the corresponding labeled edge to $H$. This step attaches each load through the highest-scoring legal connection available under the sampled node attributes.

\paragraph{Component Reconnection}

Disconnected components may remain after primary-backbone construction and load attachment. Let $\operatorname{Comp}(H)$ denote the connected components of the current graph. The set of legal inter-component bridges is
\begin{equation}
\begin{aligned}
    \mathcal{B}(H)
    =
    \bigg\{
    (u,v,k)
    \;\bigg|\;
    &u\in C_i,\;
    v\in C_j,\;
    C_i\neq C_j,\\
    &k\in
    \{
    \texttt{CONDUCTOR},
    \texttt{TRANSFORMER}
    \},\\
    &M_{uv}^{*}(k)=1,\\
    &H\cup\{(u,v,k)\}
    \text{ preserves the applicable feeder rules}
    \bigg\}.
\end{aligned}
\label{eq:legal-bridge-set}
\end{equation}

The highest-scoring bridge is
\begin{equation}
    (u^*,v^*,k^*)
    =
    \arg\max_{(u,v,k)\in\mathcal{B}(H)}
    \rho_{uv}(k).
    \label{eq:component-repair}
\end{equation}
The selected bridge is added, and the process is repeated until no disconnected component remains or no feasible bridge is available. Because every repair edge joins two distinct components, the operation cannot create a cycle within an existing component.

Algorithm~\ref{alg:pgdigress} summarizes the complete PG-DiGress sampling procedure.

\begin{algorithm}[h]
\caption{PG-DiGress Sampling}
\label{alg:pgdigress}
\begin{algorithmic}[1]
\REQUIRE Trained denoiser $\phi_\theta$, feeder rule system $\mathcal{R}$, diffusion steps $T$, and guidance schedule $\{\lambda_t\}_{t=1}^{T}$
\STATE Sample an initial noisy graph $G^T$
\FOR{$t=T,T-1,\ldots,1$}
    \STATE Predict clean node and edge distributions with $\phi_\theta(G^t,t)$
    \STATE Compute the baseline DiGress reverse probabilities $\pi_Z^t$ and $\pi_E^t$
    \STATE Decode node metadata from $\arg\max \hat{p}_{Z,\theta}^{0}$
    \STATE Construct edge legality masks $M_{uv}^t(k)$
    \STATE Reweight edge probabilities using Eq.~\eqref{eq:masked-edge-distribution}
    \STATE Sample node labels from the unmodified distribution $\pi_Z^t$
    \STATE Sample edge labels from the masked distribution $\widetilde{\pi}_E^t$
\ENDFOR
\STATE Set the preliminary graph $\widetilde{G}\leftarrow G^0$
\STATE Keep the sampled node labels $\widetilde{Z}$ fixed
\STATE Recompute the final compatibility mask $M^*$ from $\widetilde{Z}$
\STATE Construct the radial primary backbone using Eq.~\eqref{eq:primary-backbone}
\STATE Attach sampled load nodes using Eq.~\eqref{eq:load-attachment}
\STATE Reconnect remaining components using Eq.~\eqref{eq:component-repair}
\RETURN Final projected graph $G^*=(V,\widetilde{Z},E^*)$
\end{algorithmic}
\end{algorithm}

This design preserves a clear division of responsibility: the diffusion model generates node attributes and edge preferences, the soft mask improves local compatibility, and the final projection repairs global topology without changing the sampled node labels.

\section{Experiments}
\label{sec:experiments}

We design the experiments to answer four research questions:

\begin{enumerate}
    \item[\textbf{RQ1}] Does explicit rule guidance improve feeder-rule compliance?
    \item[\textbf{RQ2}] What are the respective contributions of the soft edge mask and final projection?
    \item[\textbf{RQ3}] Does rule guidance preserve the structural variation learned from the reference feeders?
    \item[\textbf{RQ4}] Can the generated topologies support downstream feeder-model construction and power-flow analysis?
\end{enumerate}

The first two questions are addressed through a controlled ablation study, while the latter two evaluate distributional fidelity and downstream usability.

\subsection{Dataset}

We use 1,019 feeder graphs derived from Synthetic Models for Advanced, Realistic Testing: Distribution Systems and Scenarios (SMART-DS)~\cite{Krishnan_Bugbee_Elgindy_Mateo_Duenas_Postigo_Lacroix_Gomez_Palmintier_2020}. Each feeder is represented as an attributed, undirected graph whose nodes carry a semantic type and phase label and whose edges carry a categorical connection class. The node-type space is
\begin{equation}
    \mathcal{X}
    =
    \{
    \texttt{SOURCE},
    \texttt{LOAD},
    \texttt{OTHER}
    \},
\end{equation}
and the edge-label space is
\begin{equation}
    \mathcal{L}_E
    =
    \{
    \texttt{NO\_EDGE},
    \texttt{CONDUCTOR},
    \texttt{TRANSFORMER}
    \}.
\end{equation}
Phase labels follow the primary and secondary categories defined in Section~\ref{sec:problem}. The explicit \texttt{NO\_EDGE} class allows the diffusion model to assign a categorical label to every candidate node pair.

Table~\ref{tab:data-summary} summarizes the node and active-edge categories in the processed dataset. 

\begin{table}[h]
\centering
\caption{Node and active-edge categories in the processed SMART-DS feeder dataset.}
\label{tab:data-summary}
\begin{tabular}{llr}
\toprule
Category & Interpretation & Count \\
\midrule
\texttt{SOURCE}
    & Feeder source node
    & 1,019 \\
\texttt{LOAD}
    & End-use or building load
    & 15,306 \\
\texttt{OTHER}
    & Intermediate bus, pole, or junction
    & 18,767 \\
\texttt{CONDUCTOR}
    & Same-voltage electrical connection
    & 30,455 \\
\texttt{TRANSFORMER}
    & Primary--secondary voltage transition
    & 3,618 \\
\bottomrule
\end{tabular}
\end{table}

We split the data at the feeder level into $80\%$ training, $10\%$ validation, and $10\%$ test sets, ensuring that no feeder appears in more than one subset. All distributional comparisons use the held-out test feeders as the reference set.

\subsection{Implementation Details}

For training, we keep only the categorical attributes modeled by PG-DiGress: node type and phase from node feature indices $0$ and $5$, and edge type from edge feature index $1$. Node type and phase are concatenated into a single categorical label, such as \texttt{LOAD-S1}, and then one-hot encoded. Edge labels are also one-hot encoded, with an additional \texttt{NO\_EDGE} class introduced during dense graph conversion to represent absent edges. 

All feeders are converted to undirected graphs before training. Continuous attributes, such as geometric length and other numerical node or edge features, are discarded. The models are implemented using \texttt{PyTorch Geometric} and trained on the high-performance computing system using one NVIDIA H100 SXM GPU. 

\subsection{Ablation Study}
We compare PG-DiGress against an unconstrained DiGress baseline and two ablated variants. Because PG-DiGress modifies sampling rather than training, all four variants use the same trained DiGress denoiser and differ only in how edge generation and final projection are performed. The graph representation, trained model checkpoint, diffusion configuration, and graph-size sampling distribution are therefore held fixed across comparisons.

\begin{itemize}
    \item \textbf{DiGress}: samples node and edge labels from the original discrete reverse distributions without feeder-rule guidance or final projection.
    
    \item \textbf{Mask only}: applies the local edge-compatibility mask during reverse diffusion without final projection.
    
    \item \textbf{Projection only}: uses unconstrained reverse diffusion followed by the final projection.
    
    \item \textbf{PG-DiGress}: combines edge masking during reverse diffusion with final projection.
\end{itemize}

This comparison isolates the contribution of local rule guidance from that of global topology repair. For each variant, we generate $N_{\mathrm{gen}}=2000$ feeder graphs. Graph sizes are sampled from the same empirical node-count distribution of
the training feeders, following the original DiGress sampling procedure.
The same model checkpoint and sampling configuration are used across all
variants.

\subsection{Evaluation Metrics}
\label{sec:evaluation-metrics}

We evaluate generated feeders along three complementary dimensions:
feeder-rule compliance, structural fidelity, and downstream feeder-model construction. Together, these metrics assess whether the generated graphs satisfy the encoded feeder rules, preserve structural variation in the reference data, and can be consumed by a physics-based downstream workflow.

\paragraph{Feeder-Rule Compliance}

Local edge compliance measures whether sampled active edges are compatible with their endpoint attributes. Let
\begin{equation}
    \mathcal{E}_{\mathrm{cond}}(G)
    =
    \left\{
    \{u,v\}\in\mathcal{E}_{\mathrm{phys}}(E)
    \mid
    E_{uv}=\texttt{CONDUCTOR}
    \right\}
\end{equation}
and define $\mathcal{E}_{\mathrm{tr}}(G)$ analogously for transformer edges. The conductor-rule compliance is
\begin{equation}
    A_{\mathrm{cond}}(G)
    =
    \frac{
        \displaystyle
        \sum_{\{u,v\}\in\mathcal{E}_{\mathrm{cond}}(G)}
        M_{uv}(\texttt{CONDUCTOR})
    }{
        \max
        \left(
        1,
        |\mathcal{E}_{\mathrm{cond}}(G)|
        \right)
    },
\end{equation}
and the transformer-rule compliance is
\begin{equation}
    A_{\mathrm{tr}}(G)
    =
    \frac{
        \displaystyle
        \sum_{\{u,v\}\in\mathcal{E}_{\mathrm{tr}}(G)}
        M_{uv}(\texttt{TRANSFORMER})
    }{
        \max
        \left(
        1,
        |\mathcal{E}_{\mathrm{tr}}(G)|
        \right)
    }.
\end{equation}

We evaluate the compliance metrics at two levels. For conductor, transformer and source-to-load path criteria, we first compute a within-graph compliance rate and then report the mean over all generated samples. Single-source compliance, connectivity, primary radiality and strict feeder pass are graph-level criteria and are reported as the percentage of generated graphs meeting the corresponding condition.

We additionally report three graph-level criteria:

\begin{itemize}
    \item \textbf{Single-source compliance}: whether the graph contains exactly one source node.

    \item \textbf{Connectivity}: whether $G_{\mathrm{phys}}(E)$ is connected.

    \item \textbf{Primary radiality}: whether the primary conductor subgraph $G_{\mathrm{pri}}$ is a tree.

\end{itemize}

For each graph, we separately report the valid source-to-load path rate as
\begin{equation}
    A_{\mathrm{path}}(G)
    =
    \frac{1}{
        \max\left(1,|V_{\mathrm{load}}|\right)
    }
    \sum_{v\in V_{\mathrm{load}}}
    \mathbf{1}
    \left[
        \Pi_G(v,s)=\{\pi_{v,s}\}
        \land
        N_{\mathrm{tr}}(\pi_{v,s})
        =
        \mathbf{1}
        [\nu_v=\texttt{SECONDARY}]
    \right].
    \label{eq:path-compliance}
\end{equation}

The source--primary condition in Eq.~\eqref{eq:source-primary-admissibility}
holds by construction through the admissible node-label space $\mathcal{Z}$
and therefore does not require a separate evaluation metric.
A strict pass requires all remaining local and global feeder rules to hold simultaneously: conductor and transformer compatibility, exactly one source, graph connectivity, primary radiality, and a valid source-to-load path for every load. We therefore define
\begin{equation}
\begin{aligned}
    I_{\mathrm{strict}}(G)
    =
    \mathbf{1}
    \big[
        &A_{\mathrm{cond}}(G)=1
        \land
        A_{\mathrm{tr}}(G)=1 \\
        &\land
        |V_{\mathrm{src}}|=1
        \land
        G_{\mathrm{phys}}(E)\text{ is connected} \\
        &\land
        G_{\mathrm{pri}}\text{ is a tree}
        \land
        A_{\mathrm{path}}(G)=1
    \big].
\end{aligned}
\label{eq:strict-pass}
\end{equation}

The strict feeder pass rate is the mean of
$I_{\mathrm{strict}}(G)$ over all generated samples.

\paragraph{Structural Fidelity}

Structural fidelity measures whether generated feeders preserve the graph characteristics of the reference dataset. We compare the empirical distributions of:

\begin{itemize}
    \item number of nodes;
    \item average degree;
    \item average shortest-path length;
    \item graph diameter;
    \item algebraic connectivity; and
    \item S-metric.
\end{itemize}

Node count characterizes feeder scale, while average degree measures graph sparsity. Average shortest-path length and diameter capture the global extent of the feeder. Algebraic connectivity summarizes connectivity through the
graph Laplacian, and the S-metric characterizes degree mixing. Metrics requiring connectivity are computed on the largest connected component so that disconnected generated samples remain represented rather than being discarded.

For each scalar statistic $f$, we report its empirical distribution and the $1$-Wasserstein distance
\begin{equation}
    W_1
    \left(
        \widehat{P}_{f,\mathrm{gen}},
        \widehat{P}_{f,\mathrm{test}}
    \right)
\end{equation}
between the generated and held-out test distributions. Lower values indicate closer agreement with the reference distribution.

\paragraph{Downstream Feeder-Model Construction.}

We evaluate whether PG-DiGress samples can progress through the downstream feeder-modeling workflow. The evaluation is sequential: generated topology is first converted into a feeder model, followed by electrical-parameter assignment, power-flow execution, and convergence testing. We report the fraction of evaluated samples successfully reaching each stage.

All downstream success rates are computed over the complete set of PG-DiGress samples used in this evaluation. The strict feeder criterion in
Eq.~\eqref{eq:strict-pass} is intentionally more restrictive than the minimum requirements of the downstream model builder; consequently, a sample may fail one strict evaluation rule while still being constructible as a downstream
feeder model.

Successful feeder-model construction establishes that the generated categorical topology can be consumed by the downstream workflow. Power-flow execution and convergence additionally depend on the assigned electrical
parameters and operating conditions. We therefore treat convergence as an operational check under the tested scenario rather than as a guarantee of feasibility under arbitrary operating conditions.

\section{Results}
\label{sec:results}

We present the results according to the four experimental questions defined
in Section~\ref{sec:experiments}. We first evaluate feeder-rule compliance
and use the ablations to separate the contributions of local masking and
global projection. We then examine whether rule guidance preserves the
structural distribution of the reference feeders, and finally test whether PG-DiGress samples can be converted into downstream feeder models and executed in power-flow analysis.

\subsection{Rule Guidance Improves Feeder Compliance}

\begin{table*}[!t]
\centering
\caption{
Feeder-rule compliance and ablation results.
Within-graph metrics show the mean fraction of compliant edges or loads
across generated samples, whereas graph-level metrics show the percentage of complete graphs satisfying each criterion.
The strict feeder pass rate requires all six conditions to hold simultaneously. Higher values are better, and the best result in each row is shown in bold.
}
\label{tab:validity}

\small
\begin{tabularx}{\textwidth}{>{\raggedright\arraybackslash}X *{4}{>{\centering\arraybackslash}p{0.13\textwidth}}}
\toprule
Metric
    & DiGress
    & Soft mask only
    & Projection only
    & PG-DiGress \\
\midrule

\rowcolor{localmetric}
\multicolumn{5}{l}{\textbf{Within-graph compliance}} \\

Mean conductor-edge compliance (\%)
    & 89.2
    & \textbf{100.0}
    & 88.4
    & \textbf{100.0} \\

Mean transformer-edge compliance (\%)
    & 54.3
    & 99.3
    & 87.5
    & \textbf{100.0} \\

Mean source-to-load path compliance (\%)
    & 14.3
    & 94.3
    & 96.0
    & \textbf{97.1} \\

\addlinespace[2pt]
\rowcolor{globalmetric}
\multicolumn{5}{l}{\textbf{Graph-level compliance}} \\

Single-source compliance (\%)
    & 73.4
    & 45.2
    & 92.3
    & \textbf{98.9} \\

Connectivity (\%)
    & 77.2
    & 42.3
    & \textbf{100.0}
    & \textbf{100.0} \\

Primary radiality (\%)
    & 83.2
    & 27.5
    & 98.5
    & \textbf{99.2} \\

\midrule
\textbf{Strict feeder pass rate (\%)}
    & 13.7
    & 27.1
    & 86.5
    & \textbf{96.8} \\

\bottomrule
\end{tabularx}
\end{table*}

Table~\ref{tab:validity} compares unconstrained DiGress, the two ablated variants, and the complete PG-DiGress method. PG-DiGress increases the strict feeder pass rate from $13.7\%$ to $96.8\%$, an improvement of $83.1$
percentage points. 

The ablations show a clear separation between local and global failure modes. The soft-mask variant primarily improves within-graph compatibility: mean conductor-edge compliance increases from $89.2\%$ to $100.0\%$, transformer-edge compliance from $54.3\%$ to $99.3\%$, and source-to-load
path compliance from $14.3\%$ to $94.3\%$. However, only $42.3\%$ of the generated graphs are connected and $27.5\%$ satisfy primary radiality, resulting in a strict feeder pass rate of $27.1\%$. Thus, local guidance alone does not impose the required global feeder structure.

The final-projection variant exhibits the complementary behavior. It achieves $100\%$ connectivity and $98.5\%$ primary radiality, but residual local edge incompatibilities remain, with mean conductor- and transformer-edge compliance of $88.4\%$ and $87.5\%$, respectively. Its strict feeder pass
rate therefore reaches $86.5\%$, substantially higher than the soft-mask variant but still below the complete method.

Combining both modules achieves strong performance at both levels: PG-DiGress achieves $100\%$ mean conductor- and transformer-edge compliance, $100\%$ connectivity, and $99.2\%$ primary radiality, with a strict feeder pass rate of $96.8\%$. These results support the intended division of responsibility in PG-DiGress: the soft mask improves local compatibility during reverse diffusion, while the final projection repairs the remaining global feeder structure.

\paragraph{Qualitative comparison.}
The qualitative example in Figure~\ref{fig:qual_exp} illustrates the same
failure mode visually. The DiGress sample has a visually plausible size and branching structure but contains two source nodes, violating the single-source requirement and making the graph unacceptable as a feeder topology. In contrast, the PG-DiGress sample contains a single source and forms a source-connected topology with conductor and transformer assignments consistent with the sampled node attributes. This example complements the aggregate results by showing a domain-specific violation that is not apparent from graph size or density alone.

\subsection{Rule Guidance Preserves Structural Variation}
\label{sec:results-fidelity}

\begin{figure*}[!tbh]
    \centering
    \includegraphics[width=0.9\textwidth]
    {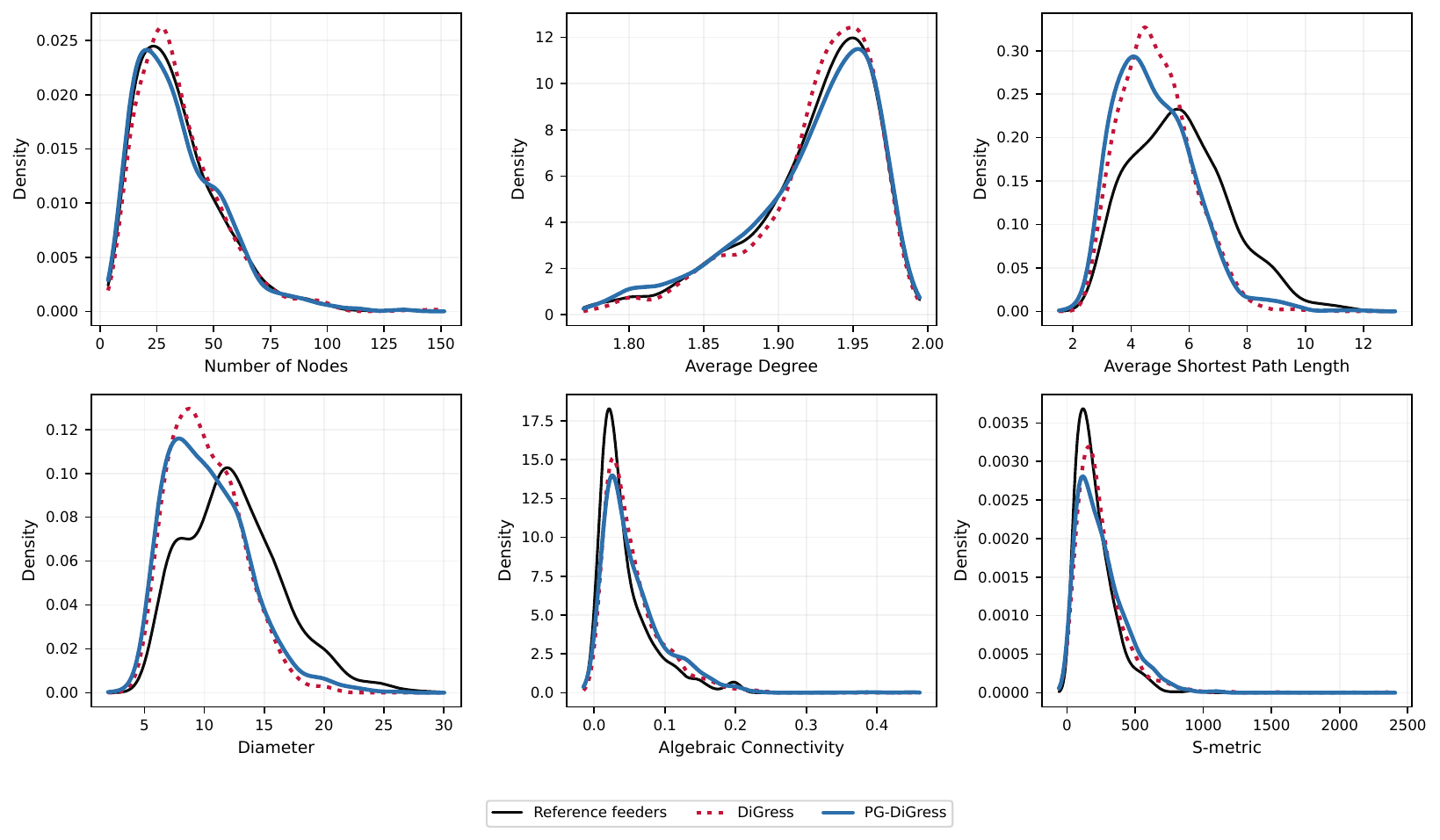}
    \Description{Distributions of feeder size, degree, path length, diameter, algebraic connectivity, and S-metric for held-out reference feeders, DiGress samples, and PG-DiGress samples.}
    \caption{
    Structural distributions of held-out reference feeders, unconstrained DiGress samples, and PG-DiGress samples. Both generative models closely match the reference distributions of feeder size and average degree. PG-DiGress remains close to DiGress across the reported statistics, indicating that rule guidance substantially improves feeder-rule compliance without sacrificing structural fidelity.
    }
    \label{fig:graph-statistics}
\end{figure*}

Figure~\ref{fig:graph-statistics} compares the structural distributions of the held-out reference feeders, unconstrained DiGress samples, and PG-DiGress samples. 

As expected, the generated node-count distributions closely match the
reference distribution because graph size is sampled from the empirical training distribution. Because each method samples graph sizes independently
from this same distribution, small differences between the generated node-count curves reflect finite-sample variation rather than differences introduced by PG-DiGress. We therefore treat node count primarily as a sampling sanity check and focus the structural comparison on properties induced by the generated connectivity.
Their average-degree distributions are also similar to the reference distribution, although this statistic is less discriminative for sparse radial graphs, whose average degree naturally approaches two.

The largest remaining discrepancies occur in the path-based statistics. Both generated distributions are shifted toward shorter average path lengths and smaller diameters, indicating that the generated feeders are generally more compact than the held-out reference topologies. Differences also remain in the tails of the algebraic-connectivity and S-metric distributions.

Despite these remaining gaps, PG-DiGress achieves a lower $1$-Wasserstein distance than unconstrained DiGress for every reported statistic in Table~\ref{tab:structural-distance}. For example, the distance decreases from $0.846$ to $0.663$ for average shortest-path length, from $2.350$ to $1.264$ for diameter, and from $40.457$ to $32.503$ for the S-metric. Thus, the substantial improvement in
feeder-rule compliance does not come at the cost of structural fidelity. Across the reported statistics, PG-DiGress modestly improves agreement with the reference distribution.

\begin{table}[!t]
\centering
\caption{
$1$-Wasserstein distance between generated and held-out reference feeder distributions. Lower values $\downarrow$ indicate better structural fidelity.
}
\label{tab:structural-distance}
\begin{tabular}{lrr}
\toprule
Statistic & DiGress & PG-DiGress \\
\midrule
Number of nodes
    & 0.660 & \textbf{0.480} \\
Average degree
    & 0.003 & \textbf{0.001} \\
Average shortest-path length
    & 0.846 & \textbf{0.663} \\
Diameter
    & 2.350 & \textbf{1.264} \\
Algebraic connectivity
    & 0.009 & \textbf{0.007} \\
S-metric
    & 40.457 & \textbf{32.503} \\
\bottomrule
\end{tabular}
\end{table}
\FloatBarrier
\subsection{Downstream Feeder-Model Construction}

We finally test whether PG-DiGress topologies can progress from generated graphs to executable feeder models. This evaluation is sequential: a topology must first support feeder-model construction and electrical-parameter assignment before power-flow execution can be attempted. Table~\ref{tab:downstream} reports the fraction of evaluated PG-DiGress samples successfully reaching each stage.

\begin{table}[hb]
\centering
\caption{
Fraction of evaluated PG-DiGress samples successfully reaching each
stage of the downstream feeder-modeling workflow.
}
\label{tab:downstream}
\begin{tabular}{lr}
\toprule
Downstream stage & PG-DiGress success rate (\%) \\
\midrule
Feeder-model construction & 100 \\
Electrical-parameter assignment & 100 \\
Power-flow execution & 99.8 \\
Power-flow convergence & 99.3 \\
\bottomrule
\end{tabular}
\end{table}

All downstream rates are computed over the complete PG-DiGress evaluation set. The strict feeder criterion in Eq.~\eqref{eq:strict-pass} is more restrictive than the minimum requirements of the downstream model builder; therefore, a small number of samples that fail one strict evaluation rule can nevertheless be instantiated as feeder models.

Unconstrained DiGress is evaluated comparatively through the feeder-rule and structural metrics above, but is not used for this end-to-end test because its samples do not reliably satisfy the structural prerequisites required by the downstream construction workflow. In contrast, all evaluated PG-DiGress samples can be converted into feeder models and assigned the required electrical parameters; $99.8\%$ proceed to power-flow execution and $99.3\%$ converge successfully. 

These results connect graph-generation quality to practical downstream use. Rule compliance alone establishes that the generated categorical topology is consistent with the feeder representation, whereas successful model construction shows that the topology can be consumed by a physics-based workflow. Power-flow convergence provides a further operational check, but does not by itself establish feasibility under all loading, voltage, thermal, protection, or contingency conditions.

\section{Discussion}
\label{sec:discussion}

PG-DiGress shows that feeder knowledge can be introduced into discrete graph diffusion without redesigning the training objective. The soft edge mask and final projection address different levels of the problem: the mask guides conductor and transformer sampling toward locally compatible assignments, while the projection resolves connectivity, radiality, and source-to-load
structure after denoising. Importantly, the projection uses learned edge probabilities to rank admissible connections, rather than replacing learned generation with a fully rule-based construction procedure.

\paragraph{Rule compliance and structural fidelity.}

Feeder-rule compliance and structural fidelity measure complementary aspects of generation. Generic graph statistics describe agreement with the reference distribution but cannot detect violations such as incompatible edge classes, multiple sources, or invalid source-to-load paths. Conversely, enforcing feeder rules may alter graph-level structure through edge repair. Both forms of evaluation are therefore necessary.

In our experiments, this trade-off is limited: PG-DiGress substantially improves the strict feeder pass rate while achieving lower $1$-Wasserstein distances than DiGress across all reported structural statistics. Rule guidance therefore improves feeder compliance without collapsing the learned variation, although the generated feeders remain more compact than the SMART-DS references in path length and diameter.

\paragraph{Scope and limitations.}

PG-DiGress modifies only edge generation. Node types and phase labels remain sampled from the learned distribution, so the current method cannot directly repair invalid source counts or inconsistent node attributes. The encoded rules also establish categorical and topological consistency rather than full operational feasibility. Voltage quality, thermal loading, losses, phase balance, protection behavior, and contingency response require electrical
parameter assignment and explicit simulation.

The learned distribution is further limited by the SMART-DS feeders and the categorical representation used in this study. Evaluation across additional regions, feeder sizes, utilities, and data-construction procedures is needed to assess transferability beyond this dataset.

\paragraph{Future directions.}

Future work can extend rule guidance to node attributes, condition generation on partial topology, geography, loads, or existing infrastructure, and incorporate power-flow or resilience feedback into sampling. These extensions would support feeder completion and move the method from rule-compliant topology generation toward conditional and operationally informed distribution-system synthesis.

\section{Conclusion}
\label{sec:conclusion}

We introduced Power-Grid-constrained DiGress (PG-DiGress), a constraint-guided discrete diffusion framework for attributed
distribution-feeder topology generation. PG-DiGress follows a minimal-intervention design: it retains the standard DiGress training process, guides local edge sampling through a soft compatibility mask, and applies a projection to construct the final global feeder structure.

On SMART-DS feeder graphs, PG-DiGress increases the strict feeder pass rate from $13.7\%$ to $96.8\%$ while achieving lower structural-distribution distances than unconstrained DiGress across all reported statistics. The
generated topologies also support downstream feeder-model construction and power-flow execution, connecting improved graph generation to practical power-system analysis.

The current method improves edge compatibility and topology conditional on the sampled node attributes, but it does not guarantee valid node labels or operational power-flow feasibility. Extending rule guidance to node generation and incorporating electrical simulation feedback are important steps toward conditional and operationally informed distribution-system synthesis.

\begin{acks}
This work was authored by the National Laboratory of the Rockies for the U.S. Department of Energy (DOE), operated under Contract No. DE-AC36-08GO28308. This work was supported by the Laboratory Directed Research and Development (LDRD) Program at the National Laboratory of the Rockies. This research was performed using computational resources sponsored by the U.S. Department of Energy's Office of Critical Minerals and Energy Innovation and located at the National Laboratory of the Rockies. The views expressed in the article do not necessarily represent the views of the DOE or the U.S. Government. The U.S. Government retains and the publisher, by accepting the article for publication, acknowledges that the U.S. Government retains a nonexclusive, paid-up, irrevocable, worldwide license to publish or reproduce the published form of this work, or allow others to do so, for U.S. Government purposes.
\end{acks}

\bibliographystyle{ACM-Reference-Format}
\bibliography{full/references_ACM}

\end{document}